\documentclass[runningheads]{llncs}
\usepackage[T1]{fontenc}
 
\usepackage[utf8]{inputenc}
 
\usepackage{times}
\usepackage{url}
\usepackage{latexsym}
\usepackage{fancyhdr}
\usepackage{graphicx} 
\usepackage{CJKutf8}
\usepackage{tabularx}
\usepackage{multirow}

\usepackage{multicol} % 用于多列布局
\usepackage{graphicx}
\begin{document}
\begin{CJK*}{UTF8}{gbsn}
\title{AuditWen: An Open-Source Large Language Model for Audit}
\author{Jiajia Huang\inst{1} \and
Haoran Zhu\inst{1} \and
Chao Xu\inst{1} \and Tianming Zhan\inst{1} \and
Qianqian Xie \inst{2} \and Jimin Huang \inst{2}}
\authorrunning{J. Huang et al.}
\institute{Nanjing Audit University, 211815, Nanjing, China 
	\email{\{huangjj, xuchao, ztm\}@nau.edu.cn, zhuhr@stu.nau.edu.cn} \\ 
%\and Nanjing Audit University, 211815, Nanjing, China 
%	\email{ zhuhr@stu.nau.edu.cn } \\
\and The FinAI, Singapore 
  \email{xqq.sincere@gmail.com, jimin@chancefocus.com} }
%\and The FinAI, Singapore 
 % \email{jimin@chancefocus.com} }
%
\maketitle              % typeset the header of the contribution
\begin{abstract}
Intelligent auditing represents a crucial advancement in modern audit practices, enhancing both the quality and efficiency of audits within the realm of artificial intelligence. With the rise of large language model (LLM), there is enormous potential for intelligent models to contribute to audit domain. However, general LLMs applied in audit domain face the challenges of lacking specialized knowledge and the presence of data biases. To overcome these challenges, this study introduces AuditWen, an open-source audit LLM by fine-tuning Qwen with constructing instruction data from audit domain. We first outline the application scenarios for LLMs in the audit and extract requirements that shape the development of LLMs tailored for audit purposes. We then propose an audit LLM, called AuditWen, by fine-tuning Qwen with constructing 30k instruction dataset from 15 audit tasks and 3 layers. In evaluation stage, we proposed a benchmark with 5k instructions that covers a set of critical audit tasks derived from the application scenarios. With the benchmark, we compare AuditWen with other existing LLMs from information extraction, question answering and document generation. The experimental results demonstrate superior performance of AuditWen both in question understanding and answer generation, making it an immediately valuable tool for audit.

\end{abstract}
\section{Introduction}

Audit is an independent economic supervision activity conducted by governmental agencies or a special organ in accordance with the law to conduct pre-and-post-event reviews of major projects and financial revenues and expenditures of financial institutions or enterprises. In recent years, with the development of big data, the data foundation and audit methodology of national audit are also undergoing changes \cite{Zhang1: 2020}. The audit methodology is transitioning from big data audit to intelligent audit \cite{Huang: 2023}, aiming at recommending or selecting the optimal strategy for audit decision-making through the extensive integration of machine learning, deep learning, and other information technologies.

With the emergence of ChatGPT \footnote{https://chat.openai.com}, large language models (LLMs) \cite{Che: 2023} have attracted much attention from researchers. Its smooth natural dialogue and document generation capabilities have rendered it widely used in various fields, such as in  financial \cite{Xie: 2023}, medical \cite{Singhal: 2023}, legal \cite{Dai: 2023} and so on. A large language model is a deep learning model with a very high number of parameters and computational power that can automatically learn the syntax, semantics, and context of input natural language and can generate text of corresponding to it. As a powerful artificial intelligence technology, large language model possess a strong capacity for understanding and generating natural language and can provide innovative solutions for the audit.

However, the current general LLMs commonly encounter issues like a deficiency in domain-specific knowledge and the existence of data bias. Similar to their application in other domain-oriented tasks, LLMs face challenges when directly applied to auditing, including difficulties in understanding input issues clearly and providing accurate responses to fact-based tasks, a phenomenon known as hallucination \cite{Che: 2023}. Moreover, auditors argue that intelligent auditing with LLMs should prioritize collaboration between individuals and the model to jointly accomplish complex audit tasks \cite{Huang: 2023}. This demand necessitates that LLMs not only comprehend concepts, entities, and knowledge within the audit domain, but also master the fundamental processes of audit work to assist auditors in achieving high-quality results. LLMs excel in context memory, knowledge retrieval, and text generation, thereby offering unique advantages in this regard. 

Therefore, it is essential to train a LLM specifically for the audit domain, aligning with the actual requirements and raw data of auditing practices. By refining and tailoring LLM tasks to align with auditing requirements, the audit-focused LLM should grasp the terminology, concepts, and regulations of auditing, ultimately delivering more precise and dependable results, especially for the complicated audit tasks. Guided by the practical applications of national audit, this study aims to identify potential uses of LLM in the audit domain, collect high-quality audit-relevant raw texts and further construct an instruction dataset to build a large language model  tailored for audit by fine-tuning a state-of-the-art LLM. This model is referred to as AuditWen.   

The contributions of this study are as follows:

{\bf(1) Scenarios abstraction}. We have categorized the application scenarios of LLM in audit as core requirements, regulatory requirements, and derived requirements. The abstracted scenarios can serve as a roadmap for future researchers to advance the development of LLMs for auditing purposes.

{\bf(2) Multi-audit-tasks}. We abstract the corresponding NLP (natural language processing) tasks of LLM from 3 layers, including (a) phrase layer with information extraction and phrase classification, (b) sentence layer with audit-issue summary, audit legal recommendation and QA tasks, (c) document layer with audit risk analysis and audit report generation. 

{\bf(3) First open-source audit LLM}. It is the first open-source LLM for audit. We have openly released the AuditWen \footnote{The AuditWen is available at : https://github.com/HooRin/AuditWen}, including the instruction dataset, the evaluation benchmark and the model to encourage open research and transparency in the research field.

{\bf(4) Outstanding performance}. AuditWen shows significant performance on various of audit NLP tasks compared with the state-of-the-art LLMs, especially in audit issue summary and legal recommendation. AuditWen can be directly used in some audit practice scenario.

\section{Related Works}

{\bf Open Sourced Large Language Models}.
The GPT (Generative Pre-Training) series of models released by OpenAI has ushered in a new era of large language model. GPTs and other LLMs demonstrate powerful language understanding and generation capabilities through pre-training on extensive text datasets followed by fine-tuning for diverse NLP tasks. Most of the open-source LLMs, such as LLaMA \cite{Touvron: 2023}, Alpaca \cite{Taori: 2023} , Baichuan \cite{Yang: 2023}, ChatGLM\footnote{https://github.com/THUDM/ChatGLM-6B}, Qwen-VL Chat \cite{Bai: 2023},  have parameters ranging from 7B and 13B up to 65B. This rapid increase in the number of parameters results in notable enhancements in model power and performance, enabling LLMs to excel in NLP tasks. Generally, LLM building process consists of four main stages, i.e., pre-training, supervised fine-tuning (SFT), reward modeling and reinforcement learning from human feedback.Among the four stages, supervised fine-tuning of a base LLM with instruction dataset can produce superior answers to user queries compared to the base model, all at a lower cost. Along this line, some domain LLMs are proposed by constructing domain-oriented instruction dataset and fine-tuning base LLM (e.g,. LLaMA) with the dataset. For example, PIXIU \cite{Xie: 2023}  is an LLM specialized in financial domain, whereas HuaTuo \cite{Wang: 2023} is tailored for the medical domain, both fine-tuned using LLaMA. However, there is currently a lack of open-source LLMs and instruction tuning data specifically tailored for auditing purposes.

{\bf LLM tasks and domain-oriented benchmarks}. 
To compare the performance of different LLMs, researchers have designed various types of LLM evaluation benchmarks and released evaluation reports \cite{Cheng: 2023}\cite{Guo: 2023}. Among them, Microsoft Research Asia \cite{Guo: 2023} has comprehensively sorted out and summarized 219 relevant studies from the perspectives of evaluation objects, evaluation fields and evaluation methods. In general, the current evaluation tasks are mainly designed from the perspectives of information extraction, text classification and text generation. The evaluation tasks of information extraction mainly include named entity recognition (NER) and key element recognition. The task of text classification includes emotion classification, text classification and entity classification. Text generation tasks include answer generation based on input question, machine translation, document generation in a specified form. Based on the above classification of evaluation tasks, researchers have released the open-sources of the domain evaluation benchmark datasets and fine-tuned domain large language models, such as PIXIU\cite{Xie: 2023}, FinBen \cite{Xie: 2024}, LAiW \cite{Dai: 2023}, HuaTuo \cite{Wang: 2023} and so on.

Currently, there is no established benchmark for evaluating LLMs in the field of audit. According to the audit service requirements, this study designs 15 different LLM tasks across 3 layers, constructs the corresponding instruction datasets, and release multi-dimensional evaluation results for both existing mainstream LLMs and our fine-tuned audit-specific LLM, AuditWen.

\section{Application Scenarios of LLM in Audit Domain}
\subsection{Audit issue summary and laws recommendation}
The primary task of audit is to identify any potential audit issues within a project and determine which laws and regulations can serve as the audit basis. From this perspective, auditors are seeking LLMs to assist in summarizing audit issues based on audit working papers and recommending suitable laws and regulations as both qualitative and punishment basis.  

\begin{figure}
	\centering
	\includegraphics[width=0.95\linewidth]{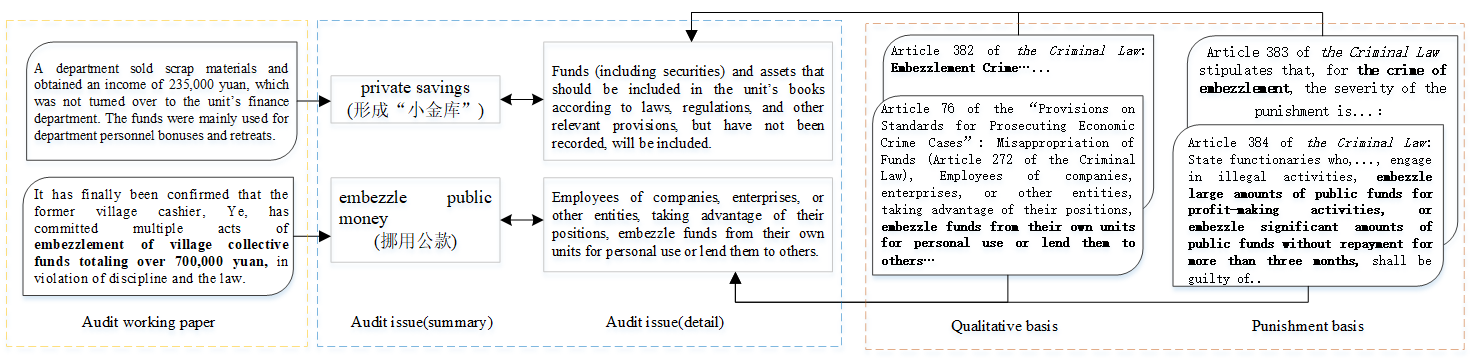}
	\caption{An audit issue schema derived from audit working paper and laws and regulations. }
	\label{fig:1}
\end{figure}

The primary challenge in the application is that an internal auditor may have a divergent qualitative basis for an audit issue compared to a social auditor based on the case description in the audit working paper. For example, an internal auditor may use items from enterprise internal control manual as qualitative basis without any penalty provision, while a social auditor may refer to items in \textit{Accounting Law and Criminal Law} for punishment.  To address this challenge, we propose an audit issue schema that summarizes audit issue from case description and aligns them with the clauses of laws and regulations simultaneously, as shown in Figure \ref{fig:1}. We hope to bridge a gap between the clause of laws and regulations and the audit issue. 

\subsection{Audit Relevant Question and Answer}
The secondary task of LLM used in audit is to answer question related to audit, such as questions list in Table \ref{tab:dummy-1}. These questions pertain to defining an audit concept, understanding the specifics of a particular clause of a law, determining the methods for investigating and verifying audit issues, and identifying the necessary data to be collected. These diverse questions prompt us to gather relevant audit documents pertaining to audit cases, audit criteria, audit guidelines, and so on. When assessing the quality of answers generated by LLM, it is crucial to minimize the occurrence of hallucination responses and ensure the retrieval of original text based on existing system documents and other relevant content. 

\begin{table}
	\centering

	{
		\begin{tabularx}{\textwidth}{
				>{\hsize=.8\hsize\linewidth=\hsize}X
				>{\hsize=1.2\hsize\linewidth=\hsize}X
			}
			\hline \hline
			\bf Query& \bf Answer \\ \hline
			What internal control information does the company need to disclose? (公司需要披露哪些内部控制信息？) &  The Company shall fully disclose any internal control information that has a significant impact on investors' investment decisions.(凡对投资者投资决策有重大影响的内部控制信息，公司均应充分披露。) \\
			\hline
			What are the responsibilities of the audit institution under the Internal Audit Regulations? (内部审计条例规定的审计机关的职责有哪些？) & According to Article 23 of Chapter 5 of the Internal Audit Regulations of XX Province, the responsibilities of audit institutions include the following:… (根据XX省内部审计条例第五章第二十三条，审计机关的职责包含以下几项：...) \\
			\hline
			What methods should be adopted in the audit of significant secured loans to identify and conduct detailed investigation? (重大担保借款审计中需要采取哪些方法来识别并展开详细调查？) &Review meeting minutes and materials, documents, contract approval documents, etc., review decision-making process, guarantee scope, scale, etc. (调阅会议纪要及上会材料、文书、合同审批文件等，审核决策流程、担保范围、规模等。) \\
			\hline
			What key information should be collected when conducting a construction project performance audit? (在进行建设项目绩效审计时，应该收集哪些关键资料？) & When conducting a construction project performance audit, key data include: approved project proposals and feasibility study reports,… (在进行建设项目绩效审计时，关键资料包括：已获批准的项目建议书和可行性研究报告，...) \\
			\hline
		\end{tabularx}
	}
		\caption{Examples of possible QA proposed by auditor.}\label{tab:dummy-1}
\end{table}

\subsection{Audit assistant}
Further derive requirement of LLM applied in audit domain is LLM can act as an intelligent assistant and help auditor to extract specified phrase from audit document, do accounting relevant numerical calculation, generate an outline for an audit report and further fill content based on the given audit working papers. The possible case questions are list in Table \ref{tab:dummy-2}. Audit assistant usually need to execute fine-grained NLP task step by step, such as information extraction, multi-documents summarization and document generation.Additionally, audit assistants must achieve collaborative work between humans and machines with the guidance of human-provided knowledge.

\begin{table}
	\centering

	{
		\begin{tabularx}{\textwidth}{
				>{\hsize=.2\hsize\linewidth=\hsize}X
				>{\hsize=1.8\hsize\linewidth=\hsize}X
			}
			\hline 	\hline
			\bf Id & \bf Query \\ \hline
			\multirow{2}*{Q1} &  Please extract entity about the audited organization from the following documents. (请从下面文档中抽取出被审计单位信息。) \\
			\hline
			\multirow{2}*{Q2} & Please judge whether Company A is losing money according to the following statement.(请根据下面的报表判断A公司是否亏损？) \\
			\hline
			\multirow{2}*{Q3} & Please write a business leader economic responsibility audit report template. (请撰写出一个企业领导人经济责任审计报告模板。)\\
			\hline
			\multirow{3}*{Q4} & Based on the uploaded audit draft and the generated audit report template, please write the audit process and method of XX leader's accountability audit. (基于上传的审计底稿和生成的审计报告模板，请撰写XX领导人经责审计的审计过程与方法。) \\
			\hline
			\multirow{3}*{Q5} & Based on the uploaded audit ledger list, audit ledgers belonging to the same audit issue are merged into the same document and output. (基于上传的审计台账列表，将属于同一审计问题的审计台账归并到同一文档中并输出。) \\
			\hline
		\end{tabularx}
	}
	\caption{The potential tasks that may be assigned to an audit intelligent assistant.}\label{tab:dummy-2}
\end{table}

\section{AIT: Audit Instruction Dataset and Tuning}

In this section, we initially outline the tasks of audit LLM based on the application scenarios of audit. Then we collect source data and design relevant instruction dataset and evaluation benchmark for audit LLM. At last, we build AuditWen by fine-tuning Qwen \cite{Bai: 2023} with AIT.

\subsection{Task abstraction for audit LLM}
Based on the application scenarios of audit, we abstract the audit tasks from three levels, namely, sentence, paragraph and documents, as example shown in Table \ref{tab:dummy-4}.
\subsubsection{Sentence level}
This level focus on information extraction from sentence and phrase classification.

{\bf Audit NER.} Accurately extract audit entity from text is the most elementary task for understanding audit content.We have developed an audit name entity recognition (NER) datasets from annotated sentences that include three types of entities, ORG, audit-issue and audit-basis, as shown in Table \ref{tab:dummy-3}.

{\bf Relation Classification.} Based on two audit entities extracted from a sentence, this task needs to predict the relation between the entity pair from given category set. The relations are defined in Table \ref{tab:dummy-7} in Appendix. This task can be used to expand audit knowledge graph by extracting information from unstructured text using LLM. 

{\bf Phrase classification.} Predict the category of an audit phrase from a set of options, where the phrase is (1) an audit-item entity that need to be classified into one of the given audit item type. (2) An audit issue relevant entity that need to be classified into one of the given audit type. (3) An law and regulation name that need to be classified into one of the given law and regulation category. 

\begin{table}
	\centering
	
	{
		\begin{tabularx}{\textwidth}{
				>{\hsize=0.2\hsize\linewidth=\hsize}X
				>{\hsize=0.4\hsize\linewidth=\hsize}X
				>{\hsize=0.4\hsize\linewidth=\hsize}X
			}
			\hline 	\hline
			\bf Entity tag & \bf Description & \bf Examples \\ \hline
			audit issue（审计问题） &  word or phrase of expressing an audit issue & 同一个人账户重复缴存,  规避招标，“小金库”\\
			\hline
			audit basis（审计依据） & word or phrase of expressing a law or regulation name & 招标投标法，中华人民共和国刑法，会计准则 \\
			\hline
			audit organization（审计对象）& entity of expressing an organization under audit &  国家机关、民办非企业单位，城市发展银行 \\
			\hline
		\end{tabularx}
	}
	\caption{Audit entity types defined in audit domain.}\label{tab:dummy-3}
\end{table}

\subsubsection{ Paragraph level}

Question answer (QA) is the task of answering an audit question based on provided information, as shown in Table \ref{tab:dummy-1}. In this level, we defined several types of question and answer tasks to make LLM understand the common question in audit.

{\bf Definition of audit entity}, namely answer the definition of an audit entity, such as\textit{ what is internal audit?} The task makes LLM understand the  concept and explanation of common audit entity. 

{\bf Audit-legal relevant question}, namely answer the question related to audit law, standards, guidelines. These part of QA pairs are very important for tuning an audit LLM, since the core scenario of audit LLM is to recommend appropriated laws and regulations as the audit basis for given audit issue.

{\bf Audit-issue relevant question}, namely answer the question related to audit issue, including (1) use a phrase to summarize the audit issue based on case description, (2) describe the specific performance of an audit issue, (3) recommend appropriate laws for a given audit issue.

{\bf Other-audit relevant question}. These QA pairs refer to (1) what method can be used in an audit case and what material need to prepare further, (2) what is the objective of an audit project, (3) list out the audit items of an audit project.

\subsubsection{Documents level}
This level focus on comprehensive documents analysis and generation, including audit risk/problem analysis, audit case/report generation, as shown in Q3-Q5 of Table \ref{tab:dummy-2}.

{\bf Risk/problem analysis}, namely analyzes the latent risks or issues of an audit project based on provided background information. 

{\bf Audit document generation}, namely generate an outline, or a template or a complete document based on input query, including (1) generate the audit process for a certain audit case, (2) outline the structure of an audit report for a specific audit matter.

\subsection{Instruction dataset construction}
Building upon the audit-oriented LLM tasks, we have developed an Audit Instruction Tuning dataset (AIT) specific to each task. Based on raw texts collected from audit domain discussed in Section 5.1, we need to construct a proper instruction for each of the raw texts. 

First of all, for sentence level tasks and part of questions presented in paragraph level, we write five different instructions for each task and evaluate their performance on current LLM based on PIXIU project \footnote{PIXIU is available at: https://github.com/chancefocus/PIXIU}.Then the best instruction is saved for further constructing more instruction data. For audit-legal relevant question in paragraph level that concerns to items in audit laws, we used GPT-4 to generate a question and corresponding answer. For audit report generation task, we write one proper instruction for it because the query of this task is concise. AIT is the first large-scale instruction-tuning and evaluation benchmark dataset for audit LLMs that condensed from audit applications. 

Generally, following the instructions proposed in PIXIU \cite{Xie: 2023}, we build instruction tuning samples with the following templates:

\begin{itemize}
	\item Template (1) : [Task prompt] with \{Context: [input text]\}, [question] with \{category\}, Answer: [output]
	\item Template (2): [Task prompt] with Context: [input text] and [question],  Answer:[output]
	
\end{itemize}

[task prompt] is the prompt designed for each type of the tasks, {category} used in classification tasks of sentence level to list out all categories, [input context] contains the input audit context of each task, such as a sentence or a paragraph. [question] is the final question or demand based on Context. [output] is the corresponding answer for the input text, such as the category in classification task or the truth answer in QA task. Examples of the instruction of each task is shown in Table \ref{tab:dummy-4}.

\begin{table}
	\centering
	
	{
		\begin{tabularx}{\textwidth}{
				>{\hsize=0.15\hsize\linewidth=\hsize}X
				>{\hsize=0.06\hsize\linewidth=\hsize}X
				>{\hsize=0.79\hsize\linewidth=\hsize}X}
			\hline \hline
			\bf {Task name} & \bf Template & \bf {Examples of an instruction data} \\
			\hline
			\multirow{8}*{Audit NER} &\multirow{8}* {T(1)} & "query": 文本: 通过对证券公司和国有企业审计，…。","在上面的文本中，请完成命名实体识别任务，即识别代表审计疑点('auditissue')、机构('ORG')、审计法律法规('auditlbasis')三类实体类型的实体名称，答案应遵循格式:"实体名称, 实体类型"。\\
			& & "answer": "证券公司, ORG".\\
			& & "label": ["O", "O", "O", "B-ORG", "I-ORG", "I-ORG", "I-ORG", "",…] \\
			\hline
			\multirow{2}*{Definition of}  & \multirow{4}*{T(2)} & "query": 请问什么是营业外收入？\\
			\multirow{2}*{audit entity} & & "answer": 该科目核算的是企业发生的与其生产经营无直接关系的各项收入，包括固定资产盘盈、处置固定资产净收益….。 \\
			\hline
			\multirow{2}*{Audit-legal}  & \multirow{4}*{T(2)} & "query": 如果某公司违反了中华人民共和国证券法第九十条关于征集股东权利的规定，将受到何种法律后果？\\
			\multirow{2}*{relevant quesiton} & & "answer":根据中华人民共和国证券法第一百九十九条，该公司将被责令改正并给予警告…。\\
			\hline
			\multirow{2}*{Risk/problem}  & \multirow{5}*{T(2)} & "query": 在国有企业经济责任审计，资产审计可能存在哪些审计风险？\\
			\multirow{3}*{analysis}& &	"answer": 资产审计可能存在如下风险点：（一）客户管理效率低，没有全面调研客户资质、信用状况并动态跟踪…。\\
			\hline
			\multicolumn{2}{c}{Audit document generation} & Referred in Table \ref{tab:dummy-8} \\
			\hline
		\end{tabularx}
	}
	\caption{Examples of the instruction data used in LLM tuning dataset.} \label{tab:dummy-4}
\end{table}

\subsection{Fine-tuning}

We further build AuditWen by fine-tuning Qwen \cite{Bai: 2023} with AIT because AIT is Chinese dataset and evaluation results on several LLMs show that Qwen achieves best performance on our evaluation benchmark dataset. To fine-tune the audit LLM, the audit instruction datasets outlined in Section 4.2 are divided into training, validation, and test sets. All the tasks in the training and validation sets are mixed together for fine-tuning, while each test set is utilized to evaluate the performance of AuditWen and other baseline LLMs.

We fine-tune Qwen-7B-chat\footnote{The model of Qwen-7B-chat is downloaded from https://huggingface.co/Qwen/Qwen-7B-Chat/tree/main} with 15 epochs based on AdamW optimizer \cite{Loshchilov and Hutter: 2017}. The batch size is set to 8, the initial learning rate is 3e-4, learning rate scheduler type choose as cosine, and warm up steps to 0.01. The AuditWen is fine-tuned on 8*A40 GPU with LoRA (Low-Rank Adaptation) \cite{Hu: 2023} where the LoRA rank set to 64, LoRA alpha set to 16 and LoRA dropout set to 0.05. The maximum length of input texts is 2048. We choose LoRA for fine-tuning is because the method can make LLM achieve a good result in downstream task with training a few additional parameters. The addition parameter matrix merges with the large-scale of original parameters by reparametrization to form a new model for inference. 

\section{Experimental Results}
\subsection{Statistics of instruction dataset}

To obtain domain data source for fine-tuning an audit LLM, we collect raw documents that relevant to definition of audit entity , audit relevant laws and kinds of structured audit cases that describe the detail process of an audit project, including audit issue, audit method, audit punish law and audit items. The raw data collected from baidubaike, public audit textbook, open law and other public website. 

From the raw dataset, we construct an entity-relation classification dataset where two audit entities extracted from a given sentence and it’s need to classify the relation between them from given category set. Here, the relations between an entity pair are defined in Table \ref{tab:dummy-7} in Appendix, and the truth category tag is labeled by human annotation. The rest of the classification tasks and entity extraction tasks are constructed with the similar way. Based on the raw classification task description and truth category tag, we converted each of them into instruction data with Template (1), as discussed in Section 4.2.

To construct audit-legal relevant instruction dataset, we gathered a substantial amount of audit-relevant laws, regulations, criterions and segmented each raw law or regulation into individual items. Then, GPT4 \cite{OpenAI: 2023} is utilized to generate a question-answer pair (QA pair) based on the input items, as the examples shown in Table \ref{tab:dummy-8} in Appendix. The similarity between the original legal-item and the generated QA-pair are evaluated by BERT Score (F1) \cite{Zhang2: 2020}. The similarity analysis reveals that over 80.1\% of QA pairs exhibit a similarity score greater than 0.8, while 19\% of QA pairs fall within the similarity range of 0.7 to 0.8,, which denotes that GTP\-4 can generate QA pair from given legal-item with high quality. Therefore, these QA-pairs can serve as instruction data that effectively capture the essence of the original legal content.  

For the audit case/report generation task, we collected some representative audit cases or reports with various forms and convert each of them into an instruction data, where the query is a short instruction while the answer is a long document with given form. For the rest of the tasks in paragraph level, raw information are extracted from structured audit cases and converted into instruction data with Template (2) in accordance with specific conditions, as the examples shown in Table \ref{tab:dummy-4}. 

All of the train, validation and test sets for each of the tasks are shown in Table 5. For audit entity classification, only a test set is created to assess the generalization capability of AuditWen on untrained tasks. Therefore, 5-shot evaluation are employed for the task. In addition, as in the audit NER task, three new types of entities are defined that have not been encountered in base LLM, we also employ 5-shot prompting for evaluation. The rest of the tasks are evaluated under zero-shot prompting.

\begin{table}
	\centering
	
	{
		\begin{tabularx}{\textwidth}{
				>{\hsize=0.10\hsize\linewidth=\hsize}X
				>{\hsize=0.15\hsize\linewidth=\hsize}X
				>{\hsize=0.35\hsize\linewidth=\hsize}X
				>{\hsize=0.2\hsize\linewidth=\hsize}X
				>{\hsize=0.2\hsize\linewidth=\hsize}X
			}
			\hline \hline
			\bf Level & \bf Task name & \bf Sub-task name & \bf \#train/val./test & \bf Annotation \\ \hline
			\multirow{5}* {\shortstack {Sentence \\level}} & \multicolumn{2}{l}{Audit NER} & 4091/1022/1424 & human annotation \\
			& \multicolumn{2}{l}{Relation classification} & 817/232/117 & human annotation \\
			&\multirow{3}*{\shortstack {Phrase \\ classification}} & audit entity cla. (AEC) & —/—/1578  &\multirow{3}*{human annotation} \\	 
			& & audit-issue phrase cla. (AIC) & 1210/344/166 & \\
			& & legal name cla. (LNC) & 1463/418/218 & \\
			\hline
			\multirow{8}* {\shortstack {Paragraph \\ level}}	& \multicolumn{2}{l} {Definition of audit entity} & 1756/500/19 &  extract from raw text\\
			&\multicolumn{2}{l} {Audit-legal relevant question} & 15774/112/505 & generated by GPT-4 \\
			& \multirow{3}*{Audit issue} & audit issue summary (AIS) & 253/71/36 & \multirow{3}* {extract from raw text} \\
			& &  audit issue describe (AID) & 202/56/29 & \\
			& &  legal  recommendation (LR) & 1567/445/224 & \\
			& \multirow{3}*{\shortstack {Other-audit \\ relevant \\ question}} & audit procedures and material (APM) & 671/190/96 &\multirow{3}*{extract from raw text}  \\
			& & audit type and objectives (ATO) & 609/171/87 & \\
			& & Other question (OQ) & 903/257/129 & \\
			& & & & \\
			\hline
			\multirow{2}*{\shortstack {Documents \\ level}} & \multicolumn{2}{c}{Audit case analysis} & 544/151/77 & \multirow{2}*{extract from raw text} \\
			& \multicolumn{2}{c}{Audit doc. generation} & 48/11/6 &  \\
			\hline
			\multicolumn{3}{c}{Total}  & \multicolumn{2}{l} {29908/3980/4941}\\
			\hline
		\end{tabularx}
	}
	\caption{The details of our evaluation datasets. "Annotation" denotes the construction manner of the instruction data from raw data. source.}\label{tab:dummy-5}
\end{table}

\subsection{Evaluation of different LLMs}

{ \bf Baseline Models}. Several strong and representative baseline models are selected to compare with our AuditWen model. For open-sources LLMs, Qwen-7B-Chat, ChatGLM3-6B are selected to perform zero-shot or 5-shot prompting on the audit evaluation benchmark dataset. For close-source LLM, GPT-4 \cite{OpenAI: 2023} is selected. 

{ \bf Evaluation Metrics}. As the tasks in sentence level are information extraction and classification, missing is employed to evaluate the  proportion of prediction results that can be successfully inferred from LLM , while accuracy and F1 are employed to evaluate the classification effectiveness. As the tasks in paragraph level and document(s) level are Q\&A task, BERT Score (F1) \cite{Zhang2: 2020}, BART Score \cite{Yuan: 2021} are employed to evaluated the similarity between the predict answer and the truth answer. For these two metrics, pre-train models with Chinese language are utilized, i.e., \textit{bert-base-chinese} and \textit{CPT} \cite{Shao: 2021}. In addition, we evaluate the definition of audit entity and legal recommendation with ROUGE \cite{Lin and Hovy: 2003}, because the answer of these tasks need to be more precise compared with other QA tasks. As word segmentation is a part of ROUGE evaluation, a user dictionary specific to the audit domain is created and loaded into the jieba segmentation tool. For the rest of the tasks, BERT Score (F1) and BART Score are used to evaluate the answer quality.

\begin{table} 
	\centering

	{
		\begin{tabularx}{\textwidth}{
				>{\hsize=0.12\hsize\linewidth=\hsize}X
				>{\hsize=0.13\hsize\linewidth=\hsize}X
				>{\hsize=0.15\hsize\linewidth=\hsize}X
				>{\hsize=0.15\hsize\linewidth=\hsize}X
				>{\hsize=0.15\hsize\linewidth=\hsize}X
				>{\hsize=0.15\hsize\linewidth=\hsize}X
				>{\hsize=0.15\hsize\linewidth=\hsize}X
			}
		 
			\hline \hline
			\bf Task name &  \bf Sub-task name & \bf Metric & \bf {Qwen-7B-Chat} & \bf ChatGLM3-6B &  \bf GPT-4 &   \bf AuditWen \\
			\hline
			\multicolumn{2}{c}{Audit NER} & Entity\_F1 & 0.140 & 0.015 &	0.108 &	\textbf{0.512} \\
			\hline
			\multicolumn{2}{c}{\multirow{3}*{Relation classification}} & Accuracy &	--/0.085*	& 0.376/0.342*&	0.402/0.624*	& \textbf{0.615}/0.188*  \\ 
			& &  F1	 &--/0.037*	&0.243/0.373*&	0.432/0.649*&	\textbf{0.744}/0.220* \\
			& & Missing &	0.410/0.00* &	0.008/0.000* &	\textbf{0.000/0.000}*	& 0.350/0.274*\\
			\hline
			\multirow{7}*{\shortstack {Phrase \\classification}} & \multirow{3}*{AEC}  & Accuracy &	0.716/0.763*&	0.493/0.540 &	0.679/\textbf{0.810}*&	0.601/0.720*  \\
			& & F1  &	0.710/0.734* &	0.583/0.612* &	0.697\textbf{/0.816}* &	0.612/0.716*\\
			& & Missing	&0.042/0.00 &	0.146/0.000 &
			\textbf{0.023/0.000}* &	0.077/0.000* \\
			
			&\multirow{3}*{AIC} & Accuracy&	--/0.399* &	0.254/0.353*&	0.464/0.543*&	0.437/\textbf{0.601}* \\
			&	&  F1&	--/0.347*&	0.193/0.252* &	0.484/0.557*	&0.428/\textbf{0.595}* \\
			& & Missing &	0.751/0.000*&	0.078/0.058*	& \textbf{0.000/0.000}*	& 0.085/0.037* \\
			&\multirow{3}*{LNC}& Accuracy &--/0.146*&	0.394/0.468*&	0.637/0.647*&	\textbf{0.752}/0.431* \\
			&	& F1&	--/0.075*&	0.388/0.428*&	0.623/0.639*	& \textbf{0.774}/0.405*\\
			& & Missing &	0.766/0.165*	&\textbf{0.000/0.000*}	&0.004/0.000*	& 0.050/0.037* \\
			\hline
			\multicolumn{2}{c} {\multirow{5}*{Definition of audit entity}} & ROUGE-1 &0.245 & 0.22&	0.202 &	\textbf{0.298 }\\  
			&  & ROUGE-2 &0.053	&0.037&	0.037&	\textbf{0.121}\\  
			&& ROUGE-L & 0.178&	0.156&	0.121&\textbf{0.237}\\  
			&& BERT\_Score & 0.678&	0.670&	0.662&\textbf{0.702}\\  
			&& BART\_Score & -4.527&	-4.535&	-4.391&	\textbf{-4.175}\\
			\hline	 
			\multicolumn{2}{c} {\multirow{2}* {Audit-legal relevant question}} & BERT\_Score &0.696&	0.671&	0.665&	\textbf{0.723}\\  
			&& BART\_Score & -3.659&	-3.356&	\textbf{-3.424}	&-3.480\\ 
			\hline
			\multirow{9}*{Audit issue} &\multirow{2}*{AIS} & BERT\_Score &0.634	&\textbf{0.644} &	0.634&	0.642\\
			& &  BART\_Score &-4.470	&-4.485	&-4.524&\textbf{	-4.456}\\
			
			&\multirow{2}*{AID} & BERT\_Score & 0.696	&0.674&	0.655&	\textbf{0.792}\\
			&&  BART\_Score &-4.048	&-3.827	&-3.996	&\textbf{-3.044}\\
			
			&\multirow{5}*{LR} & ROUGE-1 & 0.247	&0.268	&0.275&	\textbf{0.530}\\  
			&& ROUGE-2 &0.061&	0.063&	0.083&	\textbf{0.386}\\  
			&& ROUGE-L & 0.150&	0.152&	0.151&	\textbf{0.442}\\  
			&& BERT\_Score & 0.654&	0.665&	0.677&\textbf{	0.785}\\  
			&& BART\_Score & -4.799&	-4.192	&-3.661&\textbf{-3.406}\\
			\hline
			\multirow{6}*{\shortstack {Other-audit \\ relevant \\ question}} & \multirow{2}*{APM} & BERT\_Score & 0.67&	0.682&	0.694&	\textbf{0.746}\\
			& &  BART\_Score & -5.127	&-4.681	&-5.166&\textbf{	-4.514}\\
			
			& \multirow{2}*{ATO} & BERT\_Score &0.723&	0.697&	0.634&	\textbf{0.907} \\
			& &  BART\_Score & -3.794	&-3.650	&-4.069	&\textbf{-1.587}\\
			
			& \multirow{2}*{OQ} & BERT\_Score &0.704&	0.663	&0.635&	\textbf{0.900}\\
			& &  BART\_Score & -3.284	& -3.171&	-2.985	&\textbf{-1.202}\\
			\hline
			\multicolumn{2}{c} {\multirow{2}*{Audit case analysis}}  & BERT\_Score & 0.67	&0.678&	0.667&	\textbf{0.84} \\
			& &  BART\_Score & -4.854	&-3.61&	-3.291&	\textbf{-3.031}\\
			\hline
			\multicolumn{2}{c} {\multirow{2}*{Audit doc. generation}}  & BERT\_Score & 0.658	&0.668&	0.670&\textbf{	0.684}\\
			& &  BART\_Score & -5.584	&-5.003	&\textbf{-4.782}	&-5.011 \\ 
			\hline 
		\end{tabularx}
	}
		\caption{The overall performance of different LLMs on audit evaluation benchmark, - denotes inadmissible inference result, * denotes the 5-shot evaluation result.}\label{tab:dummy-6}
\end{table}

{ \bf Overall Performance}. From the 9 audit tasks evaluation results shown in Table 6, our fine-tuned model, AuditWen, significantly outperforms its base model QWen-7B-Chat and other state-of-the-art LLMs, especially in paragraph level and document level tasks. It is because fine-tuned the base LLM with domain-oriented instruction data enables the model to acquire domain-relevant knowledge, comprehend domain-specific queries, and generate outputs in the writing style typical of the audit domain.

In the NER task, AuditWen demonstrates significantly higher entity F1 scores compared to baseline models in the 5-shot evaluation, indicating that baselines struggle to accurately identify named entities when provided with five examples from each category for inference.

In phrase classification tasks, including audit entity/ audit issue and legal name classification, AuditWen achieves competitive results compared to GPT-4, and outperforms the other models in F1 and accuracy, while ChatGLM3-6B and GPT-4 achieve much lower missing rate. Furthermore, comparing the the zero-shot evaluation results of QWen-7B-Chat and AuditWen across a range of phrase classification tasks, it is observed that QWen-7B-Chat may struggle in zero-shot inference due to a high missing rate, whereas AuditWen excels in overcoming this challenge and achieves higher accuracy.

Comparing the zero-shot and 5-shot result of different models, it is evident that baseline LLMs achieve higher accuracy and lower missing rates under the 5-shot setting, whereas AuditWen demonstrates higher accuracy under the zero-shot setting for relation classification and legal name classification(LNC). It denotes the model can be used for inference without providing extra samples, which further demonstrates the superior domain-generalization capabilities of AuditWen .   

In the paragraph level and document tasks, AuditWen achieves much higher BERT Score and BART Score in legal recommendation, other-audit relevant question and risk/problem analysis. We believe that the success of AuditWen in these tasks is not only attributed to the suitable instruction template but also to the scale of the fine-tuning dataset for the task.

\textbf{Further Analysis}. We further analyze the influences of instruction template on the performance of LLMs on different tasks. For the tasks of relation classification and entity classification, varying instruction templates yield different results for Qwen-7B-Chat and ChatGLM3-6B. For instance, when the question is placed at the beginning of the template, Qwen-7B-chat exhibits poor performance on classification tasks, with accuracy rates of 0.019 for audit-issue phrase classification and 0.009 for legal name classification. ChatGLM3-6B also shows the similar result. When constructing instruction data using Template (1), ChatGLM3-6B achieves the highest accuracy and F1 score in audit entity classification, whereas it only attains an accuracy of 0.312 on the task when the question is placed at the beginning of the template. In addition, for relation classification and legal name classification(LNC), AuditWen shows better performance on zero-shot rather than on 5-shot. This may be attributed to the model's better utilization of general patterns and prior knowledge, as well as its reduced reliance on specific tasks or domains.

\subsection{Case study}
Audit-relevant document, such as audit report, is totally different from the document in general domain. Table \ref{tab:dummy-9} in Appendix compares the audit reported generated by AuditWen and GPT-4. The result demonstrates that the AuditWen can both understand the task with generating a report outline of satisfying the template and can generate more detail content of matching the outline, although only dozens of train instructions in audit case/report generation task for fine-tuning. On the contrary, base models without fine-tuning with the task fail to generate the audit report of meeting specified format.

\section{Conclusion and Discussion}
In this study, we presented AuditWen, the first audit-oriented open-source large language model. Along with the model, we also release the fine-tune model AuditWen and the evaluation benchmark dataset. Drawing from the discussion on application scenarios of LLM in audit, we have identified various audit tasks. Subsequently, we gather and construct a large-scale audit instruction dataset to fine-tune a domain-specific large language model tailored for audit tasks. The extensive evaluation results on the proposed benchmark dataset demonstrated the effectiveness of the AuditWen. 

Nevertheless, while acknowledging the positive contribution of this study, we also recognize the following limitations.\textbf{ Resource Constraints}. Due to time constraints, the scale of dataset for fine-tuning AuditWen is limited, which may not support for fine-tuning model with larger scales. \textbf{Model and Training Constrains}. We only presented the AuditWen models with 7B parameters. Due to computational and resource constraints, AuditWen models with 14B or 30B have not been released so far.

For the further work, more relevant source texts about audit cases and statute will be collected and more elaborate tasks such as audit-issue phrase extraction from clause of statute will be constructed. Based on these dataset and tasks, we devote to train a larger-scale of audit-oriented LLM. 

\section{Acknowledgements}
The work was supported by the National Science Foundation of China (NSFC, No.61802194 and No.71972102) and Research Projects in Natural Sciences at Universities in Jiangsu Province (No.23KJB520015).


\section{C EXAMPLES }
\subsection{C1. Relations defined between entity pairs and corresponding examples}
We provide the relation define between two audit entities and shows an example of entity pair extraction from a sentence and their define relation in Table 7. 

\begin{table}
	\centering
	
	{
		\begin{tabularx}{\textwidth}{
				>{\hsize=0.20\hsize\linewidth=\hsize}X
				>{\hsize=0.20\hsize\linewidth=\hsize}X
				>{\hsize=0.15\hsize\linewidth=\hsize}X
				>{\hsize=0.45\hsize\linewidth=\hsize}X
			}
			\hline \hline
			\bf {Relation name} & \bf Description & \bf {Entity pair} & \bf {Text}\\
			\hline
			\multirow{4}*{fraud\_of\_audit} & Relation between an audit item and its audit fraud &  [住房公积金归集, 同一个人账户重复缴存] & 本词条介绍了住房公积金缴纳对象在住房公积金归集方面存在主要弊错，主要包括住房公积金缴纳对象同一个人账户重复缴存的情况.\\
			\hline
			\multirow{5}*{item\_of\_audit} &	Relation between an audit instance and specific audit items &	[证券公司负债业务， 资产负债表] &	证券公司负债业务发生后，都要通过相应的会计科目反应和核算，最终表现为资产负债表上的的负债项目，达到负债的动态业务和静态业务反应相统一。\\
			\hline
			\multirow{4}*{law\_of\_audit}	& Relation between an audit issue and the corresponding law entity &	[规避招标,招标投标法]	&《招标投标法》规定：招标方不得以任何方式将应招标的项目而不招标或将必须进行招标的项目化整为零或者以其他任何方式规避招标。\\
			\hline
			\multirow{5}*{method\_of\_audit} &	Relation between an audit item and the corresponding audit method entity &	[合同履行情况审计, 检查]	& 合同履行情况审计是指对公共工程实施过程中的造价、质量、进度、安全、环境保护和水土保持等合同约定内容的执行结果进行检查。\\
			\hline
			\multirow{8}*{org\_of\_audit} &	Relation between an audit item and the corresponding audit unit &	[国家机关, 住房公积金]
			[国有企业, 住房公积金]…	& 国家机关、国有企业、城镇集体企业、外商投资企业、城镇私营企业及其他城镇企业、事业单位、民办非企业单位、社会团体（以下统称单位）及其在职职工，应当按《住房公积金管理条例》（国务院令第350号，以下简称《条例》）的规定缴存住房公积金。\\
			\hline
			\multirow{5}*{achievement\_of\_audit}	& Relation between an audit item and the corresponding audit achievement entity	& [政府预算审计, 审计报告] &	2003年审计署首次公开政府预算审计报告，政府预算审计逐步进入社会公众及媒体的视野。\\
			\hline
			\multirow{4}*{audited\_of\_org} &	Relation between an audit item and the corresponding audited unit &	[保障性安居工程跟踪审计, 哈尔滨特派办] &	在2012年城镇保障性安居工程跟踪审计过程中，哈尔滨特派办应用联网审计数据平台进行审计并取得了较好的审计成果。\\
			\hline
			\multirow{4}*{included\_domain} & Relation between an audit item and its belonging domain &	[污染减排审计, 电力行业]	& 本文介绍了持续推进电力行业污染减排审计过程和关键特点。\\
			\hline
		\end{tabularx}
	}
	\caption{Relations defined between entities.} \label{tab:dummy-7}
\end{table}

\subsection{Examples of audit-legal relevant question generated by GPT-4.}
We provide some examples of question-answer pair (QA pair) generated by GPT4 based on the input law item in Table 8. 

\begin{table} 
	\centering
	
	{
		\begin{tabularx}{\textwidth}{
				>{\hsize=0.20\hsize\linewidth=\hsize}X
				>{\hsize=0.30\hsize\linewidth=\hsize}X
				>{\hsize=0.5\hsize\linewidth=\hsize}X
			}
			\hline \hline
			\bf {Legal Name} & \bf {Content of Item} & \bf {QA generated by GPT-4}  \\
			\hline
			《第2205 号内部审计具体准则——经济责任 & 本准则适用于党政工作部门、纪检监察机关、法院、检察院、事业单位和 &	Question：《第 2205 号内部审计具体准则——经济责任审计》适用于哪些类型的单位进行经济责任审计活动？\\
			审计》(第一章第五条) & 人民团体，国有及国有资本占控股地位或主导地位的企业（含金融机构）等单位的内部审计机构、内部审计人员所从事的经济责任审计活动，其他类型单位可以参照执行。 & Answer：该准则适用于党政工作部门、纪检监察机关、法院、检察院、事业单位和人民团体，以及国有及国有资本占控股地位或主导地位的企业（包括金融机构）等单位的内部审计机构和内部审计人员所进行的经济责任审计活动。其他类型的单位也可以参照该准则执行经济责任审计。\\
			\hline
			中华人民共和国证券法(第十三章第一百九十九条)  &	违反本法第九十条的规定征集股东权利的，责令改正，给予警告，可以处五&	Question：如果某公司违反了中华人民共和国证券法第九十条关于征集股东权利的规定，将受到何种法律后果？\\
			& 十万元以下的罚款。& Answer：根据中华人民共和国证券法第一百九十九条，该公司将被责令改正并给予警告，同时可能会被处以不超过五十万元的罚款。\\
			\hline
			中华人民共和国银行业监督管理法(第四章第三十 & 银行业金融机构有违法经营、经营管理不善等情形，不予撤销将严重危害 &	Question：根据中华人民共和国银行业监督管理法，国务院银行业监督管理机构在什么情况下有权撤销银行业金融机构？\\
			九条)& 金融秩序、损害公众利益的，国务院银行业监督管理机构有权予以撤销。& Answer：根据中华人民共和国银行业监督管理法第四章第三十九条的规定，当银行业金融机构存在违法经营或经营管理不善等情形，且这些情形不予撤销将严重危害金融秩序、损害公众利益时，国务院银行业监督管理机构有权对该金融机构予以撤销。\\
			\hline
			中华人民共和国招标投标法(第二 &	招标分为公开招标和邀请招标。公开招标，是指招&	Question：中华人民共和国招标投标法规定了哪两种招标方式？ \\
			章第十条) & 标人以招标公告的方式邀请不特定的法人或者其他组织投标。邀请招标，是指招标人以投标邀请书的方式邀请特定的法人或者其他组织投标。& Answer：中华人民共和国招标投标法规定了公开招标和邀请招标两种方式。公开招标是指招标人通过发布招标公告的方式邀请不特定法人或其他组织投标；邀请招标是指招标人通过发送投标邀请书的方式邀请特定的法人或其他组织投标。\\
			\hline
		\end{tabularx}
	}
	\caption{Relations defined between entities.} \label{tab:dummy-8}
\end{table}

\subsection{Examples of audit-report generated by different LLMs.}
We present a audit report generated by AuditWen and GPT-4 respectively.  Table 9. 
\begin{table}
	\centering
	 \label{tab:dummy-9}
	{
		\begin{tabularx}{\textwidth}{
				>{\hsize=0.15\hsize\linewidth=\hsize}X
				>{\hsize=0.85\hsize\linewidth=\hsize}X
			}
			\hline \hline
			\bf {Prompt} &  Please write a template for the leadership debriefing reprot in internal audit. (请撰写一份内部审计中领导人述职报告模板. )\\
			\hline
			\multirow{7}*{\bf { \shortstack {Ground \\ truth }} } &  ***〔姓名〕同志述职报告  \\
			& 一、基本情况 
			〔说明：本部分主要反映被审计领导干部任职期限、职责范围和分管的工作。〕       \\ 
			& 二、主要业务工作开展情况 
			〔说明： 本部分主要包括：1.任期内贯彻执行党和国家有关经济方针政策和决策部署、推动所在单位可持续发展情况；2.任期内重大经济决策事项、决策过程及其执行效果；3.任期内各项工作任务目标完成情况；4.重要规章制度及内部控制制度的制定、完善和执行情况；5.任职前和任期内重大经济遗留问题及其处理情况等内容。〕 \\
			&  三、党风廉政建设情况  
			〔说明： 包括任期内履行推动所在单位党风廉政建设职责和个人遵守廉洁从业规定的情况。〕 \\
			& 四、存在的不足和需要改进的方面 。 \\
			& 五、其他需要说明的情况  \\
			& 述职人：***〔述职人本人签字〕 
			****年**月**日	\\
			\hline
			
			\multirow{9}*{\bf{AuditWen }}   & 领导人员述职报告 \\
			& 一、基本情况介绍
			（主要包括：任职时间、分工情况、分管工作等）\\
			& 二、履行职责情况 \\
			& （一）坚持政治统领，增强贯彻执行党的路线方针政策和决策部署的坚定性：
			1．（简要叙述在贯彻落实上级部门重大政策措施、公司党组（企业党委）重要决策部署方面的情况）
			2．（简要叙述如何加强党建考核，推动各级单位落实党建工作责任的情况）
			3．（简要叙述推进党内监督，落实党风廉政建设责任的情况） \\
			&（二）强化管理，切实促进公司高质量发展
			4．（简要叙述是否及时制定或修改生产经营目标，并加强目标管理的情况）
			5．（简要叙述是否建立或完善生产经营管理制度，确保制度之间相互衔接、配套协作的情况）
			6．（简要叙述领导并督促解决生产经营中存在的突出问题，保障生产经营工作顺利进行的情况）  \\
			& （三）狠抓  建设，注重队伍建设管理
			7．（简要叙述在组织培养选拔干部方面所做的事情）
			8．（简要叙述人才规划制定和人才培养方面所做的事情）
			9．（简要叙述加强团队建设，打造优秀集体的情况）
			10．（简要叙述推进企业文化建设，传播积极向上的企业文化的情况）\\
			& （四）严格遵守规章制度，树立良好作风
			11．（简要叙述是否严格遵守财经法规和规章制度的情况）
			12．（简要叙述履行“一岗双责”情况，是否既抓好经济工作又抓好廉政工作的情况）
			13．（简要叙述接受职工群众举报和批评意见的情况）\\
			&  三、自我评价
			（简要回顾在任期内的工作表现，包括成绩和需要改进之处）\\
			& 四、具体建议和要求
			（对于后续的工作提出建议和要求，例如：加强哪方面的管理工作，如何提升干部员工的积极性等）\\
			\hline
			\multirow{7}*{\bf{GPT-4}} & 内部审计组长述职报告模板
			尊敬的[公司名称]领导及审计委员会：
			我谨代表内部审计组向大家报告[报告年份]年度的工作情况，并对即将开展的工作计划进行说明。\\
			& 一、基本情况
			1. 组织结构
			- 组长： 
			- 成员： 
			- 分配职责： 
			2. 审计范围
			- 覆盖部门： 
			- 业务流程： 
			- 财务报表：\\ 
			& 二、年度工作回顾 
			1. 主要成就
			- 完成审计项目数量： 
			- 关键发现： 
			- 实施改进措施： 
			2. 风险管理
			- 识别关键风险： 
			- 风险应对措施： 
			3. 合规性审查
			- 检查遵守情况： 
			- 发现违规行为： 
			- 整改情况：
			4. 效率与效果
			- 优化流程与节约成本： 
			- 提升内控质量： 
			5. 职业发展
			- 团队培训与个人提升：
			- 资质认证： \\
			& 三、存在问题及原因 
			1. 问题概述
			2. 影响分析
			3. 原因分析\\
			& 四、改进措施与计划 
			1. 改进目标
			2. 实施计划
			- 短期措施： 
			- 中长期措施： 
			3. 监督与评估
			- 监督机制： 
			- 评估指标： \\
			& 五、下一阶段工作展望 
			1. 审计计划
			- 重点审计项目： 
			- 预计时间表： 
			2. 资源配置
			- 人力资源： 
			- 技术投入： 
			3. 创新发展
			- 新技术应用： 
			- 流程改革：\\
			&	此致
			敬礼！
			[组长签名]
			[日期]\\
			\hline
		\end{tabularx}
	}
	\caption{Audit report template generated by different LLMs.}
\end{table}

\end{CJK*}
\end{document}